\documentclass[11pt,a4paper]{article}
\usepackage[hyperref]{emnlp-ijcnlp-2019}
\usepackage{times}
\usepackage{latexsym}

\usepackage{rotating}
\usepackage{blindtext}
\usepackage{amsmath}
\usepackage{amsfonts}
\usepackage{graphicx}
\usepackage{mathtools}
\usepackage{booktabs}
\usepackage{relsize}

\usepackage[cal=cm]{mathalfa}
\usepackage[colorinlistoftodos]{todonotes}
\usepackage{makecell, multirow,}
\usepackage[export]{adjustbox}
\usepackage{url}

\aclfinalcopy 

\title{Aligning Multilingual Word Embeddings for Cross-Modal Retrieval Task}

\author{
  Alireza Mohammadshahi \\
  IDIAP Research Inst. \\
  EPFL \\
  \small{{\tt alireza.mohammadshahi@epfl.ch}} \\\And
  R\'emi Lebret \\
  EPFL \\
  \small{{\tt remi.lebret@epfl.ch}} 
  \\\And
  Karl Aberer \\
  EPFL  \\
  \small{{\tt karl.aberer@epfl.ch}} \\}

\date{}

\begin{document}
\maketitle

\begin{abstract}
In this paper, we propose a new approach to learn multimodal multilingual embeddings for matching images and their relevant captions in two languages. We combine two existing objective functions to make images and captions close in a joint embedding space while adapting the alignment of word embeddings between existing languages in our model. We show that our approach enables better generalization, achieving state-of-the-art performance in text-to-image and image-to-text retrieval task, and caption-caption similarity task.
Two multimodal multilingual datasets are used for evaluation: Multi30k with German and English captions and Microsoft-COCO with English and Japanese captions.
\end{abstract}
\section{Introduction}

In recent years, there has been a huge and significant amount of research in text and image retrieval tasks which needs the joint modeling of both modalities. Further, a large number of image-text datasets have become available \citep{elliott:16,hodosh:13,young:14,lin:14}, and several models have been proposed to generate captions for images in the dataset \cite{lu:18,bernardi:16,anderson:17,jilu:16,mao:15,rennie:16}. There has been a great amount of research in learning a joint embedding space for texts and images in order to use the model in sentence-based image search or cross-modal retrieval task \cite{frome:13,kiros:15,donahue:15,angeliki:15,socher:14,hodosh:13,karpathy:14}. 

Previous works in image-caption task and learning a joint embedding space for texts and images are mostly related to English language, however, recently there is a large amount of research in other languages due to the availability of multilingual datasets \cite{funaki:15,elliott:16,rajendran:16,miyazaki:16,specia:16,young:14,hitschler:16,yoshikawa:17}. 
The aim of these models is to map images and their captions in a single language into a joint embedding space \cite{rajendran:16,calixto:17}.

Related to our work, \newcite{gella:17} proposed a model to learn a multilingual multimodal embedding by utilizing an image as a pivot between languages of captions. While a text encoder is trained for each language in \newcite{gella:17}, we propose instead a model that learns a shared and language-independent text encoder between languages, yielding better generalization. It is generally important to adapt word embeddings for the task at hand. Our model enables tuning of word embeddings while keeping the two languages aligned during training, building a task-specific shared embedding space for existing languages. 

In this attempt, we define a new objective function that combines a pairwise ranking loss with a loss that maintains the alignment in multiple languages.  For the latter, we use the objective function proposed in \newcite{joulin:18} for learning a linear mapping between languages inspired by cross-domain similarity local scaling (CSLS) retrieval criterion~\cite{conneau:17} which obtains the state-of-the-art performance on word translation task. 

In the next sections, the proposed approach is called Aligning Multilingual Embeddings for cross-modal retrieval (AME). With experiments on two multimodal multilingual datasets, we show that AME outperforms existing models on text-image multimodal retrieval tasks. The code we used to train and evaluate the model is available at {\tt \url{https://github.com/alirezamshi/AME-CMR}}
\section{Datasets}
We use two multilingual image-caption datasets to evaluate our model, Multi30k and Microsoft COCO \cite{elliott:16,lin:14}. 

Multi30K is a dataset with 31'014 German translations of English captions and 155'070 independently collected German and English captions. In this paper, we use independently collected captions which each image contains five German and five English captions. The training set includes 29'000 images. The validation and test sets contain 1'000 images.

MS-COCO \cite{lin:14} contains 123'287 images and five English captions per image. \newcite{yoshikawa:17} proposed a model which generates Japanese descriptions for images. We divide the dataset based on \newcite{karpathy:15}. The training set contains 113'287 images. Each validation and test set contains 5'000 images.

\section{Problem Formulation}
\subsection{Model for Learning a Multilingual Multimodal Representation}
Assume image $i$ and captions $c_{\mathsmaller{X_i}}$ and $c_{\mathsmaller{Y_i}}$ are given in two languages, $X$ and $Y$ respectively. Our aim is to learn a model where the image $i$ and its captions $c_{\mathsmaller{X_i}}$ and $c_{\mathsmaller{Y_i}}$ are close in a joint embedding space of dimension $m$. AME consists of two encoders $f_i$ and $f_c$, which encode images and captions. As multilingual text encoder, we use a recurrent neural network with gated recurrent unit (GRU). For the image encoder, we use a convolutional neural network (CNN) architecture. The similarity between a caption $c$ and an image $i$ in the joint embedding space is measured with a similarity function $P(c,i)$. 
The objective function is as follows (inspired by \newcite{gella:17}):
\begin{alignat}{2}
\label{eq:loss1}
\small
\begin{split}
L_R = \sum_{(c_{\mathsmaller{S_i}},i)}\Big(\sum_{c_{\mathsmaller{S_j}}} max\big\{0,\alpha -P(c_{\mathsmaller{S_i}},i)+ P(c_{\mathsmaller{S_j}},i)\big\} \\
+\sum_{j} max\big\{0,\alpha -P(c_{\mathsmaller{S_i}},i)+P(c_{\mathsmaller{S_i}},j)\big\}\Big)
\end{split}
\end{alignat}

Where $S$ stands for both languages, and $\alpha$ is the margin. $c_{\mathsmaller{S_j}}$ and $j$ are irrelevant caption and image of the gold-standard pair $(c_{\mathsmaller{S_i}},i)$.

\subsection{Alignment Model}
\newcommand{\ra}[1]{\renewcommand{\arraystretch}{#1}}
\begin{table*}[h!]
    \footnotesize
    \begin{minipage}{1\linewidth}
	\ra{0.8}
	\centering
	\tabcolsep=0.14cm
	\begin{tabular}{lrrrrcrrrrcc}\toprule
		& \multicolumn{4}{c}{Image to Text} & \phantom{abcd}& \multicolumn{4}{c}{Text to Image} & \phantom{abcd}\\
		\cmidrule{2-5} \cmidrule{7-10}
		& R$@$1 & R$@$5 & R$@$10 & Mr && R$@$1 & R$@$5 & R$@$10 & Mr && Alignment\\
		\midrule
		\multicolumn{1}{c}{\bf{symmetric}}\\
		Parallel~\tiny{\cite{gella:17}} & 31.7 & 62.4 & 74.1 & 3 && 24.7 & 53.9 & 65.7 & 5 && -\\
		UVS~\tiny{\cite{kiros:15}} & 23.0 & 50.7 & 62.9 & 5 && 16.8 & 42.0 & 56.5 & 8 && -\\
		EmbeddingNet~\tiny{\cite{wang:18}} & 40.7 & 69.7 & 79.2 & - && 29.2 & 59.6 & 71.7 & - && -\\
		sm-LSTM~\tiny{\cite{huang:17}} & 42.5 & 71.9 & 81.5 & 2 && 30.2 & 60.4 & 72.3 & 3 && -\\
		VSE++~\tiny{\cite{faghri:18}} & \bf{43.7} & 71.9 & 82.1 & 2 && 32.3 & 60.9 & 72.1 & 3 && -\\
		Mono & 41.4 & 74.2 & 84.2 & 2 && 32.1 & 63.0 & 73.9 & 3 && - \\
		FME & 39.2 & 71.1 & 82.1 & 2 && 29.7 & 62.5 & 74.1 & 3 && 76.81\% \\
		AME & 43.5 & \bf{77.2} & \bf{85.3} & \bf{2} && \bf{34.0} & \bf{64.2} & \bf{75.4} & \bf{3} && 66.91\% \\
		\multicolumn{1}{c}{\bf{asymmetric}}\\
		Pivot~\tiny{\cite{gella:17}} & 33.8 & 62.8 & 75.2 & 3 && 26.2 & 56.4 & 68.4 & 4 && - \\
		Parallel~\tiny{\cite{gella:17}} & 31.5 & 61.4 & 74.7 & 3 && 27.1 & 56.2 & 66.9 & 4 && - \\		
		Mono & 47.7 & 77.1 & 86.9 & 2 && 35.8 & 66.6 & 76.8 & 3 && - \\
		FME & 44.9 & 76.9 & 86.4 & 2 && 34.2 & 66.1 & 77.1 & 3 && 76.81\% \\
		AME & \bf{50.5} & \bf{79.7} & \bf{88.4} & \bf{1} && \bf{38.0} & \bf{68.5} & \bf{78.4} & \bf{2} && 73.10\% \\
		\bottomrule
	    \end{tabular}
	\end{minipage}
	\caption{\label{table1} Image-caption ranking results for English (Multi30k)}
\end{table*}
\begin{table*}[t]
    \footnotesize
	\begin{minipage}{1\linewidth}
	\ra{0.8}
	\centering
	\tabcolsep=0.196cm
	\begin{tabular}{lrrrrcrrrrcc}\toprule
		& \multicolumn{4}{c}{Image to Text} & \phantom{abcd}& \multicolumn{4}{c}{Text to Image} & \phantom{abcd} \\
		\cmidrule{2-5} \cmidrule{7-10} 
		& R$@$1 & R$@$5 & R$@$10 & Mr && R$@$1 & R$@$5 & R$@$10 & Mr && Alignment\\ \midrule
		\multicolumn{1}{c}{\bf{symmetric}}\\
		Parallel~\tiny{\cite{gella:17}} & 28.2 & 57.7 & 71.3 & 4 && 20.9 & 46.9 & 59.3 & 6 && - \\
		Mono & 34.2 & 67.5 & 79.6 & 3 && 26.5 & 54.7 & 66.2 & 4 && -\\
		FME & 36.8 & 69.4 & 80.8 & 2 && 26.6 & 56.2 & 68.5 & 4 && 76.81\%\\
		AME & \bf{39.6} & \bf{72.7} & \bf{82.7} & \bf{2} && \bf{28.9} & \bf{58.0} & \bf{68.7} & \bf{4} && 66.91\%\\	
		\multicolumn{1}{c}{\bf{asymmetric}}\\
		Pivot~\tiny{\cite{gella:17}} & 28.2 & 61.9 & 73.4 & 3 && 22.5 & 49.3 & 61.7 & 6 && -\\
		Parallel~\tiny{\cite{gella:17}} & 30.2 & 60.4 & 72.8 & 3 && 21.8 & 50.5 & 62.3 & 5 && -\\
		Mono & \bf{42.0} & 72.5 & 83.0 & 2 && 29.6 & 58.4 & 69.6 & 4 && -\\
		FME & 40.5 & 73.3 & 83.4 & 2 && 29.6 & 59.2 & \bf{72.1} & 3 && 76.81\%\\
		AME & 40.5 & \bf{74.3} & \bf{83.4} & \bf{2} && \bf{31.0} & \bf{60.5} & 70.6 & \bf{3} && 73.10\%\\
		\bottomrule
	\end{tabular}
	\end{minipage} 
	\caption{\label{table2} Image-caption ranking results for German (Multi30k)}
\end{table*}
\label{sec:alignment}
Each word $k$ in the language $X$ is defined by a word embedding $\mathrm{x}_k \in \mathbb{R}^d$ ($\mathrm{y}_k \in \mathbb{R}^d$ in the language $Y$ respectively). Given a bilingual lexicon of $N$ pairs of words, we assume the first $n$ pairs $\{(\mathrm{x}_i,\mathrm{y}_i)\}_{i=1}^n$ are the initial seeds, and our aim is to augment it to all word pairs that are not in the initial lexicons. \newcite{mikolov:13} proposed a model to learn a linear mapping $\boldsymbol{W} \in \mathbb{R}^{d \times d}$ between the source and target languages:
\begin{align}
\label{eq:loss3}
\begin{split}
& min_{\boldsymbol{W} \in \mathbb{R}^{d \times d}}\frac{1}{n}\sum_{i=1}^{n}\ell(\boldsymbol{W}x_i, y_i |x_i,y_i) \\
& \ell(\boldsymbol{W}x_i, y_i |x_i,y_i) = (\boldsymbol{W}x_i - y_i)^2
\end{split}
\end{align}

Where $\ell$ is a square loss. One can find the translation of a source word in the target language by performing a nearest neighbor search with Euclidean distance. But, the model suffers from a "hubness problem": some word embeddings become uncommonly the nearest neighbors of a great number of other words \cite{george:98,dinu:14}. 

In order to resolve this issue, \newcite{joulin:18} proposed a new objective function inspired by CSLS criterion to learn the linear mapping:
\begin{align}
\label{eq:loss2}
\small
\begin{split}
L_A = &\frac{1}{n}\sum_{i=1}^{n}-2\mathrm{x}_i^T\boldsymbol{W}^T\mathrm{y}_i+ \frac{1}{k}\sum_{\mathrm{y}_j \in \mathcal{N}_Y(\boldsymbol{W}\mathrm{x}_i)}\mathrm{x}_i^T\boldsymbol{W}^T\mathrm{y}_j\\&+\frac{1}{k}\sum_{\boldsymbol{W}\mathrm{x}_j \in \mathcal{N}_X(\mathrm{y}_i)}\mathrm{x}_j^T\boldsymbol{W}^T\mathrm{y}_i
\end{split}
\end{align}

Where $\mathcal{N}_X(\mathrm{y}_i)$ means the $k$-nearest neighbors of $\mathrm{y}_i$ in the set of source language $X$. They constrained the linear mapping $\boldsymbol{W}$ to be orthogonal, and word vectors are $l2$-normalized. 

The whole loss function is the equally weighted summation of the aforementioned objective functions:
\begin{alignat}{2}
\label{eq:loss4}
\begin{split}
L_{total} = L_R + L_A
\end{split}
\end{alignat} 

\begin{figure}
  \centering
  \includegraphics[width=\linewidth,height=5cm]{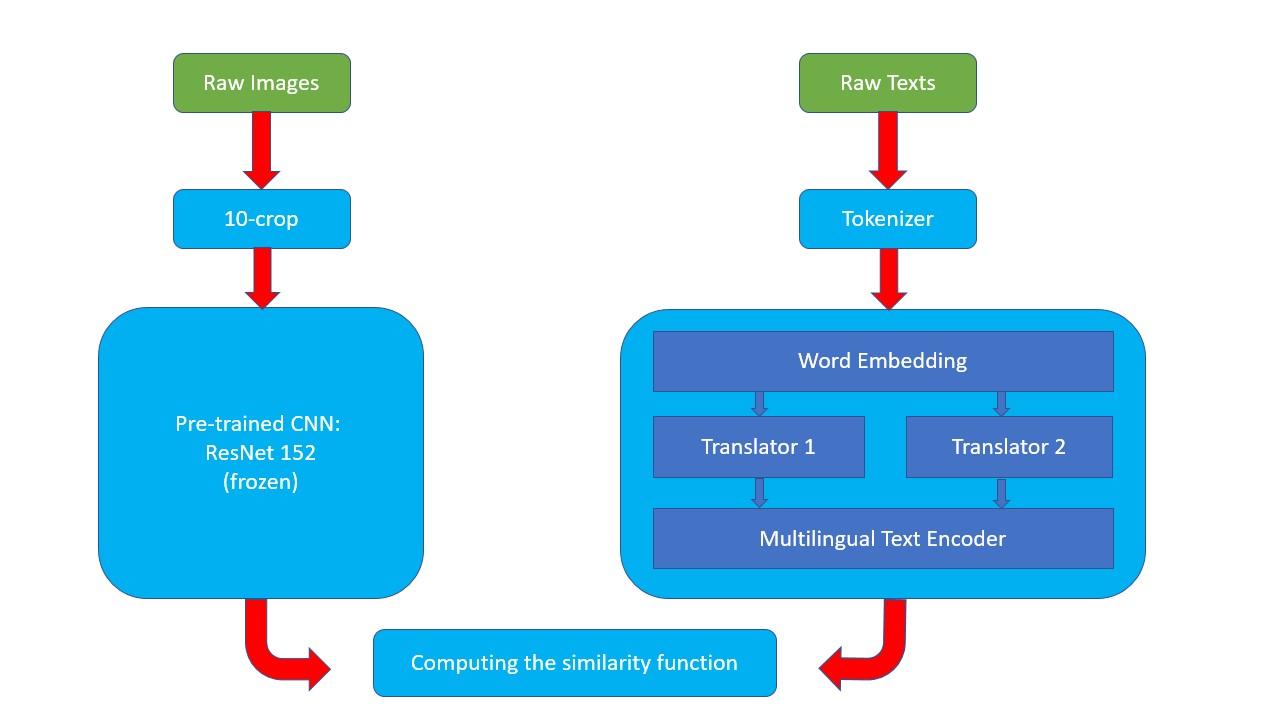}
  \caption{The AME - model architecture}
  \label{fig1}
\end{figure}

The model architecture is illustrated in Figure \ref{fig1}. We observe that updating the parameters in \eqref{eq:loss2} every $T$ iterations with learning rate $lr_{align}$ obtains the best performance. 

We use two different similarity functions, symmetric and asymmetric. For the former, we use the cosine similarity function and for the latter, we use the metric proposed in \newcite{vendrov:16}, which encodes the partial order structure of the visual-semantic hierarchy. The metric similarity is defined as:
\begin{align}
\label{eq:asym}
\begin{split}
S(a,b) = -||max(0,b-a)||^2
\end{split}
\end{align}

Where $a$ and $b$ are the embeddings of image and caption. 

\section{Experiment and Results}

\subsection{Details of Implementation \footnote{In this section, the hyper-parameters in parentheses are related to the model trained on MS-COCO.}}
We use a mini-batch of size 128. We use Adam optimizer with learning rate 0.00011 (0.00006) and with early stopping on the validation set. We set the dimensionality of joint embedding space and the GRU hidden layer to  $m=1024$. We utilize the pre-trained aligned word vectors of FastText for the initial word embeddings. For Japanese word embedding, we use pre-trained word vectors of FastText\footnote{Available at \url{https://fasttext.cc/docs/en/crawl-vectors.html}, and \url{https://fasttext.cc/docs/en/aligned-vectors.html}.}, then align it to the English word embedding with the same hyper-parameters used for MS-COCO. We set the margin $\alpha=0.2$ and $\alpha=0.05$ for symmetric and asymmetric similarity functions respectively. 

We assign $k$-nearest neighbors to be 5 (4). We set $T=500$, and $lr_{align} = 2$ $(5)$. We tokenize English and German captions with Europarl tokenizer \cite{koehn:05}. For the Japanese caption, we use Mecab analyzer~\cite{kudo:04}. We train the model for 30 (20) epochs with updating the learning rate (divided by 10) on epoch 15 (10). 

To extract features of images, we use a ResNet152 \cite{he:15} CNN architecture pre-trained on Imagenet and extract the image features from FC7, the penultimate fully connected layer. We use average features from 10-crop of the re-scaled images.
\begin{table*}[!t]
    \footnotesize
	\begin{minipage}{1\linewidth}
	\ra{0.8}
	\centering
	\tabcolsep=0.14cm
	\begin{tabular}{lrrrrcrrrrcc}\toprule
		& \multicolumn{4}{c}{Image to Text} & \phantom{abcd}& \multicolumn{4}{c}{Text to Image} & \phantom{abcd} \\
		\cmidrule{2-5} \cmidrule{7-10} 
		& R$@$1 & R$@$5 & R$@$10 & Mr && R$@$1 & R$@$5 & R$@$10 & Mr && Alignment\\ \midrule
		\multicolumn{1}{c}{\bf{symmetric}}\\
		UVS~\tiny{\cite{kiros:15}} & 43.4 & 75.7 & 85.8 & 2 && 31.0 & 66.7 & 79.9 & 3 && - \\
		EmbeddingNet~\tiny{\cite{wang:18}} & 50.4 & 79.3 & 89.4 & - && 39.8 & 75.3 & 86.6 & - && -\\
		sm-LSTM~\tiny{\cite{huang:17}} & 53.2 & 83.1 & 91.5 & 1 && 40.7 & 75.8 & 87.4 & 2 && - \\
		VSE++~\tiny{\cite{faghri:18}} & \bf{58.3} & \bf{86.1} & 93.3 & 1 && \bf{43.6} & 77.6 & 87.8 & 2 && -\\
		Mono & 51.8 & 84.8 & 93.5 & 1 && 40.0 & 77.3 & 89.4 & 2 && -\\
		FME & 42.2 & 76.6 & 91.1 & 2 && 31.2 & 69.2 & 83.7 & 3 && 92.70\%\\
		AME & 54.6 & 85 & \bf{94.3} & \bf{1} && 42.1 & \bf{78.7} & \bf{90.3} & \bf{2} && 82.54\%\\	
		\multicolumn{1}{c}{\bf{asymmetric}}\\
		Mono & 53.2 & 87.0 & 94.7 & 1 && 42.3 & 78.9 & 90 & 2 && -\\
		FME & 48.3 & 83.6 & 93.6 & 2 && 37.2 & 75.4 & 88.4 & 2 && 92.70\%\\
		AME & \bf{58.8} & \bf{88.6} & \bf{96.2} & \bf{1} && \bf{46.2} & \bf{82.5} & \bf{91.9} & \bf{2} && 84.99\%\\
		\bottomrule
	\end{tabular}
	\end{minipage} 
	\caption{\label{table3} Image-caption ranking results for English (MS-COCO)}
\end{table*}
\begin{table*}[t!]
    \footnotesize
	\begin{minipage}{1\linewidth}
	\ra{0.8}
	\centering
	\tabcolsep=0.224cm
	\begin{tabular}{lrrrrcrrrrcc}\toprule
		& \multicolumn{4}{c}{Image to Text} & \phantom{abcd}& \multicolumn{4}{c}{Text to Image} & \phantom{abcd} \\
		\cmidrule{2-5} \cmidrule{7-10} 
		& R$@$1 & R$@$5 & R$@$10 & Mr && R$@$1 & R$@$5 & R$@$10 & Mr && Alignment\\ \midrule
		\multicolumn{1}{c}{\bf{symmetric}}\\
		Mono & 42.7 & 77.7 & 88.5 & 2 && 33.1 & 69.8 & 84.3 & 3 && -\\
		FME & 40.7 & 77.7 & 88.3 & 2 && 30.0 & 68.9 & 83.1 & 3 && 92.70\%\\
		AME & \bf{50.2} & \bf{85.6} & \bf{93.1} & \bf{1} && \bf{40.2} & \bf{76.7} & \bf{87.8} & \bf{2} && 82.54\%\\	
		\multicolumn{1}{c}{\bf{asymmetric}}\\
		Mono & 49.9 & 83.4 & 93.7 & 2 && 39.7 & 76.5 & 88.3 & \bf{2} && -\\
		FME & 48.8 & 81.9 & 91.9 & 2 && 37.0 & 74.8 & 87.0 & \bf{2} && 92.70\%\\
		AME & \bf{55.5} & \bf{87.9} & \bf{95.2} & \bf{1} && \bf{44.9} & \bf{80.7} & \bf{89.3} & \bf{2} && 84.99\%\\
		\bottomrule
	\end{tabular}
	\end{minipage} 
	\caption{\label{table4} Image-caption ranking results for Japanese (MS-COCO)}
\end{table*}
\begin{table}[t]
    \footnotesize
    \centering
	\ra{0.8}
	\tabcolsep=0.055cm
	\begin{tabular}{lrrrrcrrrrc}\toprule
		& \multicolumn{3}{c}{EN $\rightarrow$ DE} & \phantom{abc}& \multicolumn{3}{c}{DE $\rightarrow$ EN} & \phantom{abc} \\
		\cmidrule{2-4} \cmidrule{6-8} 
		& R$@$1 & R$@$5 & R$@$10 && R$@$1 & R$@$5 & R$@$10\\ \midrule
		FME & 51.4 & 76.4 & 84.5 && 46.9 & 71.2 & 79.1\\
		AME & \bf{51.7} & \bf{76.7} & \bf{85.1} && \bf{49.1} & \bf{72.6} & \bf{80.5}\\
		\bottomrule
	\end{tabular}
	\caption{\label{table5} Textual similarity scores (asymmetric, Multi30k).}
    \vspace{-0.5cm}
\end{table}

For the metric of alignment, we use bilingual lexicons of Multilingual Unsupervised and Supervised Embeddings (MUSE) benchmark \cite{lample:17}. MUSE is a large-scale high-quality bilingual dictionaries for training and evaluating the translation task. We extract the training words of descriptions in two languages. For training, we combine "full" and "test" sections of MUSE, then filter them to the training words. For evaluation, we filter "train" section of MUSE to the training words. \footnote{You can find the code for building bilingual lexicons on the Github link.}

For evaluating the benefit of the proposed objective function, we compare AME with monolingual training (Mono), and multilingual training without the alignment model described in Section~\ref{sec:alignment}. For the latter, the pre-aligned word embeddings are frozen during training (FME). We add Mono since the proposed model in \newcite{gella:17} did not utilize pre-trained word embeddings for the initialization, and the image encoder is different (ResNet152 vs. VGG19). 

We compare models based on two retrieval metrics, recall at position k (R@k) and Median of ranks (Mr). 

\subsection{Multi30k Results}
In Table \ref{table1} and \ref{table2}, we show the results for English and German captions. For English captions, we see 21.28\% improvement on average compared to \newcite{kiros:15}. There is a 1.8\% boost on average compared to Mono due to more training data and multilingual text encoder. AME performs better than FME model on both symmetric and asymmetric modes, which shows the advantage of fine-tuning word embeddings during training. We have 25.26\% boost on average compared to \newcite{kiros:15} in asymmetric mode. 

For German descriptions, The results are 11.05\% better on average compared to \cite{gella:17} in symmetric mode. AME also achieves competitive or better results than FME model in German descriptions too.

\subsection{MS-COCO Results\footnote{To compare with baselines, scores are measured by averaging 5 folds of 1K test images.}}

In Table \ref{table3} and \ref{table4}, we show the performance of AME and baselines for English and Japanese captions. We achieve 10.42\% improvement on average compared to \newcite{kiros:15} in the symmetric manner. 
We show that adapting the word embedding for the task at hand, boosts the general performance, since AME model significantly outperforms FME model in both languages. 

For the Japanese captions, AME reaches 6.25\% and 3.66\% better results on average compared to monolingual model in symmetric and asymmetric modes, respectively. 

\subsection{Alignment results}

In Tables \ref{table1} and \ref{table2}, we can see that the alignment ratio for AME is 6.80\% lower than FME which means that the translators can almost keep languages aligned in Multi30k dataset.
In MS-COCO dataset, the alignment ratio for AME is 8.93\% lower compared to FME. 

\begin{figure}
  \includegraphics[width=\linewidth,height=4cm]{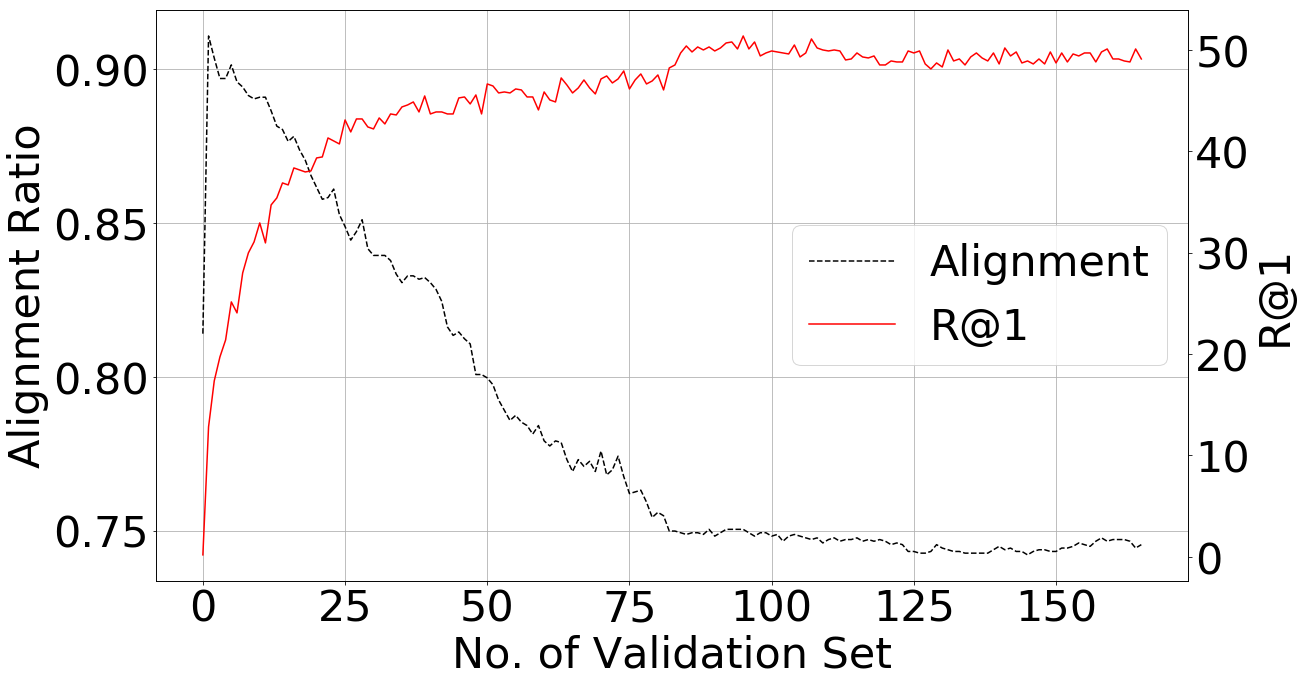}
  \caption{Alignment ratio in each validation step (asymmetric mode - image-to-text - Multi30k dataset)}
  \label{fig:alignment}
\end{figure}

We compute the alignment ratio and recall at position 1 (R@1) in each validation step. Figure \ref{fig:alignment} shows the trade-off between alignment and retrieval tasks. At the first few epochs, the model improves the alignment ratio since the retrieval task hasn't seen enough number of instances. Then, the retrieval task tries to fine-tune word embeddings. Finally, they reach an agreement near the half of training process. At this point, we update the learning rate of retrieval task to improve the performance, and the alignment ratio preserves constant. 

Additionally, we also train AME model without adding the alignment objective function, and the model breaks the alignment between the initial aligned word embeddings, so it's essential to add the alignment objective function to the retrieval task. 

\subsection{Caption-Caption Similarity Scores}
Given the caption in a language, the task is to retrieve the related caption in another language. In Table \ref{table5}, we show the performance on Multi30k dataset in asymmetric mode. AME outperforms the FME model, confirming the importance of word embeddings adaptation. 
\section{Conclusion}

We proposed a multimodal model with a shared multilingual text encoder by adapting the alignment between languages for image-description retrieval task while training. We introduced a loss function which is a combination of a pairwise ranking loss and a loss that maintains the alignment of word embeddings in multiple languages.
Through experiments with different multimodal multilingual datasets, we have shown that our approach yields better generalization performance on image-to-text and text-to-image retrieval tasks, as well as caption-caption similarity task.

In the future work, we can investigate on applying self-attention models like Transformer~\cite{vaswani2017attention} on the shared text encoder to find a more comprehensive representation for descriptions in the dataset. Additionally, we can explore the effect of a weighted summation of two loss functions instead of equally summing them together.

\section{Acknowledgement}
We gratefully acknowledge the support of NVIDIA Corporation with the donation of the Titan Xp GPU used for this research.

\newpage
\bibliography{emnlp-ijcnlp-2019.bib}
\bibliographystyle{acl_natbib}

\end{document}